\documentclass[letterpaper, 10 pt, journal, twoside]{IEEEtran}

\usepackage{amsmath,amsfonts,amssymb}
\usepackage{array}
\usepackage{booktabs}
\usepackage{graphicx}
\usepackage{textcomp}
\usepackage{stfloats}
\usepackage{url}
\usepackage{verbatim}
\usepackage{cite}
\usepackage{latexml}

\newif\ifblind
\blindfalse

\newcommand{\coderelease}{Code, trained weights and the semantic lookup table are
available at \url{https://github.com/lucashoefs/sam_vtp}, and the robot deployment
stack at \url{https://github.com/lucashoefs/sam_vtp_jetacker}.}

\begin{document}
\bstctlcite{IEEEtranBSTCTL:paper}

\title{SAM-VTP: Video Memory and Measured Geometry for Continuous Traversability Segmentation}

\ifblind
\author{Anonymous Authors%
\thanks{Author names and affiliations withheld for double-blind review.}%
}
\else
\iflatexml
\author{Lucas Hoefs \and Hui Xie \and Senjian An \and Fangyi Zhang%
\thanks{School of EECMS, Curtin University, Bentley, WA 6012, Australia.
\texttt{lucas.hoefs@postgrad.curtin.edu.au}}}
\else
\author{Lucas Hoefs$^{1}$, Hui Xie$^{1}$, Senjian An$^{1}$, and Fangyi Zhang$^{1}$%
\thanks{$^{1}$All authors are with the School of EECMS, Curtin University, Bentley, WA 6012
{\tt\footnotesize lucas.hoefs@postgrad.curtin.edu.au}}%
}
\fi
\fi

\markboth{}{}

\maketitle

\begin{abstract}
Traversability prediction is a core component of autonomous navigation, yet most vision-based methods operate frame-wise and produce temporally inconsistent predictions. We present SAM-VTP (Video Traversability Prediction), which fine-tunes SAM2's video memory pathway for dense, prompt-free traversability segmentation, conditioning each frame on a persistent memory of prior observations. Rather than inferring terrain geometry from appearance, we measure it: slope, roughness and height are computed from depth and fused into the image features. We also introduce an offline continuous supervision pipeline that generates dense per-pixel traversability targets in [0,1] from depth-derived geometry and semantic priors, requiring no robot deployment and no per-pixel annotation, capturing conditional feasibility that binary labels cannot represent. On TartanGround, a large-scale synthetic ground-robot dataset, SAM-VTP improves fuzzy Jaccard from 0.908 to 0.924 and binary IoU from 0.854 to 0.886 over SAM-TP, a frame-wise SAM2 baseline, against a seed-to-seed spread below 0.001, and cuts frame-to-frame flicker by 31\%. In closed-loop trials at two field sites, driving the same planner and controller, the baseline's dropouts leave the planner without a feasible path 3.3x more often on a paved course and 6.9x on rough ground, while SAM-VTP runs at 21.4 FPS on the Orin Nano the robot carries.
\end{abstract}

\iflatexml
\else
\fi

\section{Introduction}
\label{sec:intro}

\IEEEPARstart{T}{raversability} prediction is the task of estimating which regions of a scene are safe and feasible to traverse and is a core component of autonomous mobile robot navigation. While traversability can be inferred from LiDAR or proprioceptive signals, this work focuses on vision-based estimation from onboard camera RGB and depth imagery, formulated as a dense prediction problem. Each pixel is assigned a continuous score, and the resulting traversability map can be projected into a bird's-eye-view costmap for downstream path planning \cite{wang2025geniegeneralizablenavigationinthewild, zhang2025dualbevnavduallayerbevbased, frey2024roadrunnerlearningtraversability, mattamala2024wildvisualnavigationfast}. Traversability prediction is particularly critical in off-road, disaster-response, and cluttered indoor environments, where prior maps are unavailable or rapidly changing, and the robot must infer safe traversal directly from onboard perception \cite{annurev:/content/journals/10.1146/annurev-control-062722-100728, sevastopoulos2022surveytraversabilityestimationmobile, mattamala2024wildvisualnavigationfast}.

\begin{figure}[!t]
    \centering
    \iflatexml\includegraphics[width=\linewidth]{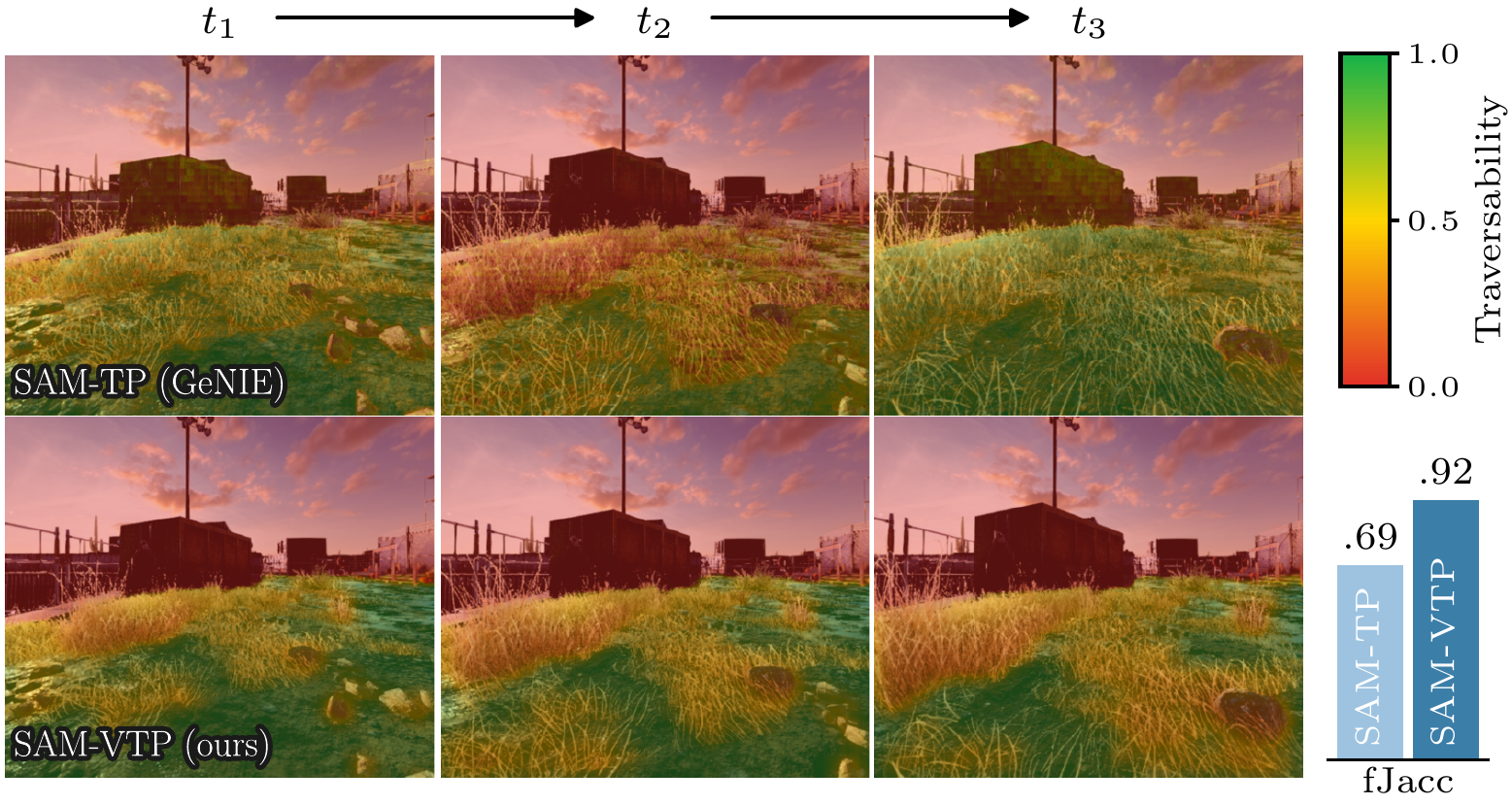}%
    \else\includegraphics[width=1.0\linewidth]{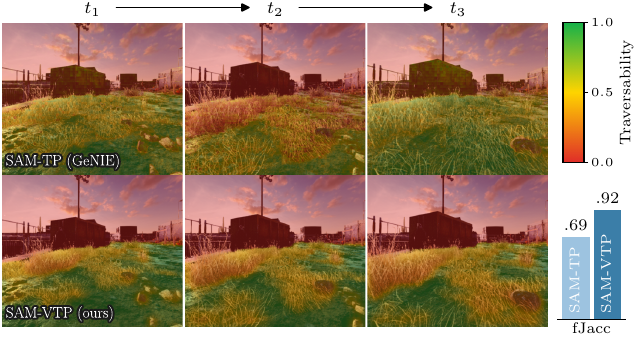}\fi
    \caption{Temporal comparison of SAM-TP (top), the frame-wise SAM2 traversability model released with GeNIE~\cite{wang2025geniegeneralizablenavigationinthewild} and the closest published system to ours, against SAM-VTP (bottom) on consecutive frames from TartanGround ConstructionSite. At $t_2$ SAM-TP loses roughly half of the traversable surface and recovers it at $t_3$, while SAM-VTP holds a coherent region across all three frames (fuzzy Jaccard, fJacc, $0.69$ against $0.92$ on this window). SAM-VTP's output is also traversability-scaled, whereas SAM-TP's is dominated by binary confidence.}
    \label{fig:overview}
\end{figure}

Foundation segmentation models encode strong priors about object boundaries and scene structure, improving generalisation across environments \cite{kirillov2023segment, ravi2024sam2}. Methods that build on them inherit those priors but not any notion of time: most existing traversability segmentation and classification methods operate in a frame-wise manner \cite{palazzo2020domainadaptationoutdoorrobot, Vecchio2024, jung2024vstrongvisualselfsupervisedtraversability, sahu2025anytraverseoffroadtraversabilityframework, patel2024roadrunnermmlearning, mattamala2024wildvisualnavigationfast, wang2025geniegeneralizablenavigationinthewild}, producing predictions independently for each image. This can produce temporally inconsistent predictions under viewpoint changes and occlusions \cite{BaghbaderaniLW024}, which lead to unstable costmaps for downstream planning; Fig.~\ref{fig:overview} shows the effect on a frame-wise baseline, whose decision on the vegetated ground flips between traversable and non-traversable across three consecutive frames of an essentially static scene. Temporal context is most often reintroduced either by recurrent models that carry a fixed-size state forward one step at a time \cite{shaban2021semantic, Gasparino_2022, gasparino2024wayfasterselfsupervisedtraversabilityprediction}, or by downstream fusion that accumulates per-frame predictions into a costmap and reconciles them with filtering or optimisation \cite{frey2024roadrunnerlearningtraversability, zhang2025dualbevnavduallayerbevbased, guan2022tnsterraintraversabilitymapping}; others reason over time at the control level \cite{zhang2024traversabilityawareleggednavigationlearning, s23063239}. Recurrent approaches compress previous observations into a fixed-dimensional state, while downstream fusion operates on already predicted maps. Neither mechanism provides the dense predictor with direct access to previously encoded visual features during the current prediction. Multi-frame attention has been applied to LiDAR traversability over pose-warped scans \cite{chen2023fastc}, and SAM2's video memory offline, to propagate dataset annotations \cite{capezzuto2026cat}. We instead place a learned visual memory inside the image-space dense predictor itself, maintained and queried at inference: each frame attends over the encoded features and mask outputs of preceding frames.

A second limitation is that terrain geometry must be inferred from RGB appearance. Surfaces that look alike can differ sharply in shape: a flat grass surface and a grassy slope are near-identical in RGB and differ only in slope. Slope, roughness and height are geometric quantities, and depth measures them, at no additional sensing cost on a stereo or RGB-D platform. We encode this measured geometry and fuse it into the image encoder's appearance features.

Additionally, traversability supervision is predominantly binary, whether it is annotated offline \cite{wang2025geniegeneralizablenavigationinthewild} or derived online from robot experience \cite{mattamala2024wildvisualnavigationfast}. In practice, traversability is conditional: a steep slope, a patch of loose gravel, or a low obstacle may be traversable for one platform and not another. Binary labels cannot encode this uncertainty and tie the supervision signal to one platform's capabilities; a graded target defers that choice to deployment. While the value of continuous traversability scores has been acknowledged in prior work \cite{guan2022tnsterraintraversabilitymapping}, no offline supervised pipeline for dense continuous per-pixel traversability supervision has been proposed. Self-supervised approaches sidestep annotation by deriving traversability from proprioceptive signals during live robot deployment~\cite{mattamala2024wildvisualnavigationfast, gasparino2024wayfasterselfsupervisedtraversabilityprediction}, but require physical interaction with the environment.

In this work, we address these limitations in three ways:
\begin{IEEEitemize}
\item A temporally-aware traversability segmentation framework built upon the video memory components of SAM2~\cite{ravi2024sam2}, enabling persistent visual reasoning across sequences of frames.
\item Measured terrain geometry: rather than leaving terrain shape to be inferred from appearance, slope, roughness and height are computed from depth, encoded, and fused into the image encoder's appearance features.
\item A continuous traversability supervision pipeline that constructs dense per-pixel targets in $[0,1]$ from depth-derived geometry and semantic priors, entirely offline and without robot deployment.
\end{IEEEitemize}
We evaluate offline and in closed loop on a physical robot, and CUDA Graph-compatible sequential inference sustains 30 FPS on a Jetson AGX Orin and 21 FPS on an Orin Nano. \coderelease

\section{Related Work}
\label{sec:relatedwork}

Early learning-based approaches replaced manual heuristics with convolutional networks trained end-to-end on RGB or RGB-D inputs, improving robustness to appearance variation across outdoor and off-road environments \cite{Giusti2016AML}. Self-supervised and weakly supervised formulations reduced annotation requirements using robot interaction signals \cite{Vecchio2024, mattamala2024wildvisualnavigationfast, Gasparino_2022, gasparino2024wayfasterselfsupervisedtraversabilityprediction, jung2024vstrongvisualselfsupervisedtraversability, jeon2024followfootprintsselfsupervisedtraversability, castro2023doesfeelselfsupervisedcostmap}. Such methods remain sensitive to domain shift, often requiring environment-specific data \cite{sivaprakasam2024salonselfsupervisedadaptivelearning, Rafi2024, chen2022calicoarsetofinealignmentsbased, Vecchio2024}, and mostly rely on hard binary or categorical supervision. Other approaches avoid task-specific training by prompting vision-language models zero-shot, relying on an operator for terrain the prompts do not cover \cite{sahu2025anytraverseoffroadtraversabilityframework}, which leaves the output as coarse as the prompt vocabulary and uncalibrated. While some works derive continuous scores from proprioceptive signals during deployment~\cite{mattamala2024wildvisualnavigationfast, gasparino2024wayfasterselfsupervisedtraversabilityprediction}, and others emit a continuous score from binary-labelled training and score it at a fixed threshold~\cite{hwang2026vita}, dense per-pixel continuous \emph{targets} generated offline have not been explored for traversability segmentation.

Terrain shape has long been scored directly. Geometric pipelines build height or elevation maps and score them by slope and roughness \cite{PAPADAKIS20131373, guan2022tnsterraintraversabilitymapping}, and learned methods increasingly combine the two, whether by predicting traversability in a bird's-eye-view grid lifted from RGB-D \cite{gasparino2024wayfasterselfsupervisedtraversabilityprediction, frey2024roadrunnerlearningtraversability}, fusing geometric with visual cues \cite{jeon2024followfootprintsselfsupervisedtraversability}, or gating between model-based and learned experts \cite{he2025navmoehybridmodellearningbased}. Image-space dense predictors, by contrast, generally leave geometry implicit in appearance features, so slope and roughness must be recovered from texture that does not determine them.

More recently, large-scale vision models such as CLIP \cite{radford2021learningtransferablevisualmodels}, DINOv2 \cite{oquab2024dinov2learningrobustvisual} and SAM \cite{kirillov2023segment} have improved visual generalisation in navigation, supplying transferable representations and segmentation priors that encode scene structure and object boundaries. Building on these representations, methods such as Wild Visual Navigation (WVN) \cite{mattamala2024wildvisualnavigationfast} and GeNIE (SAM-TP) \cite{wang2025geniegeneralizablenavigationinthewild} demonstrate improved robustness in unseen environments. Related approaches, including NaVMoE \cite{he2025navmoehybridmodellearningbased}, WayFASTER \cite{gasparino2024wayfasterselfsupervisedtraversabilityprediction}, and GFN/FSM \cite{jeon2024followfootprintsselfsupervisedtraversability}, further explore efficiency and scalability, and ViTA \cite{hwang2026vita} adapts SAM2 with learned prompt tokens and depth distilled from a monocular teacher at training time. However, despite improved generalisation, these methods predominantly operate frame-wise, producing predictions independently for each image.

Consequently, predictions can vary substantially between consecutive frames when the viewpoint changes, regions become occluded, or dynamic objects enter the scene. Recurrent modules such as ConvLSTMs \cite{gasparino2024wayfasterselfsupervisedtraversabilityprediction} compress temporal context into a fixed-dimensional latent state that is continuously overwritten, and so struggle to retain long-range spatial detail. Explicit memory architectures instead allow persistent storage and selective retrieval across time \cite{ravi2024sam2}, yet remain largely unexplored for traversability segmentation.
\section{Method}
\label{sec:method}

\subsection{Architecture}

\begin{figure}[ht]
    \centering
    \iflatexml\includegraphics[width=\linewidth]{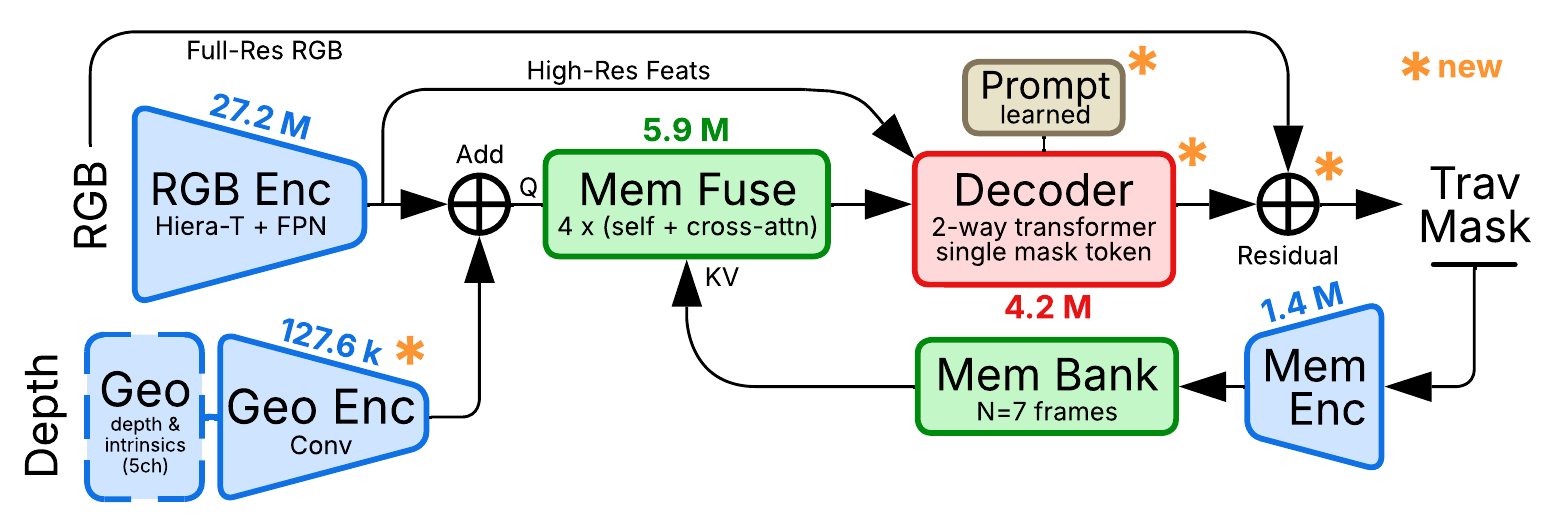}%
    \else\includegraphics[width=\linewidth]{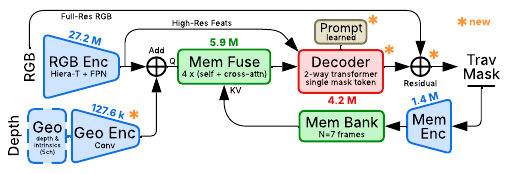}\fi
    \caption{SAM-VTP: SAM2's Hiera-T encoder and memory pathway. The four starred blocks are new or modified in this work: a depth-derived geometry encoder summed into the image features, a learned prompt replacing SAM2's prompt encoder, a decoder emitting a single mask token, and a residual head refining the output at full resolution. Parameter counts are per block; the memory bank holds activations, not parameters.}
    \label{fig:sam-vtp}
\end{figure}

We consider two SAM2-based traversability models. \textbf{SAM-TP} (from GeNIE~\cite{wang2025geniegeneralizablenavigationinthewild}) applies SAM2's image encoder and mask decoder frame-wise, skipping the video memory pathway entirely. To address the temporal inconsistency issue in SAM-TP, we propose \textbf{SAM-VTP} (Segment-Anything-Model Video-Traversability-Predictor), which retains the same decoder and prompt formulation but enables and fine-tunes SAM2's video memory pathway across contiguous sequences (Fig.~\ref{fig:sam-vtp}). A further adaptation follows from the deployment setting: terrain geometry is measured from depth rather than inferred from appearance (\ref{sssec:geometry}). The geometry input is also given to SAM-TP in the held-out ablation of Sec.~\ref{sec:results}, which resolves the asymmetry present in Table~\ref{tab:tartanground_overall}; the closed-loop experiment instead contrasts RGB-only SAM-TP with RGB-D SAM-VTP, so it compares the two systems and does not isolate memory from depth. SAM-VTP takes no pose or odometry input at any stage, so the perception stack carries no localisation dependency.

\subsubsection{Prompt-Free Decoder}
\label{subsec:prompt}

Both SAM-VTP and the SAM-TP baseline follow GeNIE~\cite{wang2025geniegeneralizablenavigationinthewild} in removing the user from SAM2's prompt interface: traversability needs no user-supplied point, since the region of interest is always the ground the robot can drive on. The prompt is therefore a constant of the task, and we learn it once during training rather than place it by hand. SAM2's prompt encoder is replaced by four learned sparse tokens together with a learned dense embedding and decoder position encoding, which condition the mask decoder in place of encoded clicks; the decoder itself is unchanged. Concurrent work adopts learned prompt tokens for traversability on the same backbone~\cite{hwang2026vita}.

\subsubsection{SAM-VTP: Video Traversability Predictor}

SAM-VTP adapts SAM2's video segmentation pipeline for traversability prediction (see Fig. \ref{fig:sam-vtp}). Given a sequence of RGB frames $\{I_t\}_{t=1}^{T}$, SAM2's Hiera image encoder encodes each frame to produce hierarchical visual features. The memory attention module conditions each frame's features on encoded representations stored in a fixed-size memory bank $\mathcal{M}$ of the $N$ most recent frames ($N=7$, matching SAM2). The decoder produces a per-pixel traversability logit map, and the memory encoder updates $\mathcal{M}$ with the current frame's features and predicted mask. The memory attention, memory encoder, and decoder are fine-tuned. The image encoder is unfrozen and optimised with a reduced learning rate (default: 0.1× the base rate) to preserve pretrained priors.

Two modifications adapt the decoder to continuous targets. First, SAM2 emits three candidate mask tokens and selects among them using a pretrained mask-quality head that receives no traversability supervision; its ranking is therefore arbitrary under our objective and free to change between frames, causing flicker unrelated to the memory pathway. We disable multi-mask output and supervise a single token. Second, the decoder predicts at stride 4 of the backbone embedding ($128{\times}128$ for a $512{\times}512$ input), so structures thinner than roughly five input pixels cannot be represented. Kerbs, rails and thin obstacles frequently carry traversability boundaries, a limitation at exactly the pixels that matter most. We therefore append a residual head that refines the upsampled logit map conditioned on the full-resolution frame.

\subsubsection{Measured Terrain Geometry}
\label{sssec:geometry}

Section~\ref{subsec:labels} defines traversability as a semantic prior modulated by terrain geometry, so a model given only RGB must recover that geometry from appearance. We instead measure it: from the depth map we derive slope, local height roughness, height, log range and validity, and add their encoding to the coarsest backbone level, feeding both memory attention and the decoder.

Because the same geometric quantities enter both the supervision targets and the model input, we take three precautions against circularity. The semantic prior carries most of the signal and is never supplied, so it must still be inferred from the image. Slope, step height and roughness are the standard criteria for ground-vehicle traversability~\cite{sevastopoulos2022surveytraversabilityestimationmobile, annurev:/content/journals/10.1146/annurev-control-062722-100728}, not a convention introduced here. The branch is also dropped for a random subset of training samples. This regularises the fusion but does not make depth optional: withholding it at inference costs $0.061$ fJacc (Section~\ref{sec:results}), so the depth input is a requirement of the deployed system rather than an auxiliary cue.

\subsubsection{Training Objective}

Continuous targets constrain which region losses are applicable. Against them the soft Jaccard loss is not minimised by predicting the target~\cite{wang2023jaccardmetriclosses}: it is a ratio, so each pixel's gradient depends on the intersection and union accumulated over the rest of the frame, and pushes that pixel to $1$ or $0$ as its target falls above or below a threshold set by that accumulation. Where targets are high, that is over-prediction of traversability, the unsafe direction. Soft Dice, soft Tversky and focal Tversky inherit the defect~\cite{wang2023dicesemimetriclosses}. We therefore adopt the Jaccard metric loss $\Delta_{\mathrm{JML},1}$~\cite{wang2023jaccardmetriclosses}, which on $[0,1]$ is exactly $1-J$ for the Ruzicka (fuzzy Jaccard) index $J$, the continuous-valued generalisation of IoU,
\begin{equation}
\label{eq:ruzicka}
J(p,g) = \frac{\sum_i \min(p_i, g_i)}{\sum_i \max(p_i, g_i)},
\qquad
\mathcal{L}_{\mathrm{Ruz}} = 1 - J(p,g),
\end{equation}
where $p_i\in[0,1]$ and $g_i\in[0,1]$ are the predicted probability and target at pixel $i$, and sums run over all pixels of a frame. $\mathcal{L}_{\mathrm{Ruz}}$ attains its optimum if and only if $p_i=g_i$ everywhere, $\Delta_{\mathrm{JML},1}$ being a metric on $[0,1]^p$~\cite{wang2023jaccardmetriclosses}, and is reduced per sample so that frames with little traversable area still contribute full-strength gradients. IoU-family losses are known to cost calibration, an effect soft targets mitigate~\cite{wang2023jaccardmetriclosses}; ours are soft everywhere.

The full objective adds \textbf{BCE} for per-pixel calibration, \textbf{Focal loss} \cite{lin2018focallossdenseobject} ($\gamma{=}3$) to emphasise hard pixels such as thin obstacles and boundaries, and a \textbf{temporal consistency} term:
\begin{equation}
\mathcal{L}
= \lambda_r \mathcal{L}_{\mathrm{Ruz}}
+ \lambda_b \mathcal{L}_{\mathrm{BCE}}
+ \lambda_f \mathcal{L}_{\mathrm{Focal}}
+ \lambda_t \mathcal{L}_{\mathrm{Temp}},
\end{equation}
where $\lambda_r$, $\lambda_b$, $\lambda_f$ and $\lambda_t$ are scalar weights on the region, BCE, focal and temporal terms, set to $(1.2,\,0.8,\,0.7,\,0.5)$ by a coarse search on the validation split. Writing $p_t=\sigma(\ell_t)$ for the predicted probabilities at frame $t$, with $\sigma$ the logistic function, the temporal term is
\begin{subequations}
\begin{gather}
\mathcal{L}_{\mathrm{Temp}}
= \frac{1}{T-1}\sum_{t=2}^{T}
e^{-\beta m_t}\,
\| p_t - p_{t-1} \|_1, \\
m_t = \frac{1}{HW}\| \bar{x}_t - \bar{x}_{t-1} \|_1, \quad
\bar{x}_t = \tfrac{1}{3}\sum_{c=1}^{3} x_t^{(c)},
\end{gather}
\end{subequations}
where $x_t^{(c)}$ is channel $c$ of frame $t$ with intensities normalised to $[0,1]$, $\bar{x}_t$ the resulting grayscale frame, $H{\times}W$ the frame resolution, and $T$ the clip length. The scalar $m_t$ measures inter-frame appearance change and acts as a motion proxy: the penalty is strong when the scene is static and relaxes as $m_t$ grows, so genuinely rapid changes are not over-penalised. The decay rate is fixed at $\beta=5$ against the unit intensity scale, giving a weight of $0.57$ at a typical $m_t\approx0.11$. A grayscale difference is a coarse motion proxy, responding to illumination change and dynamic objects as readily as to ego-motion; a flow-based measure would gate the term more precisely at additional cost.

\begin{figure}[tb]
    \centering
    \iflatexml\includegraphics[width=\linewidth]{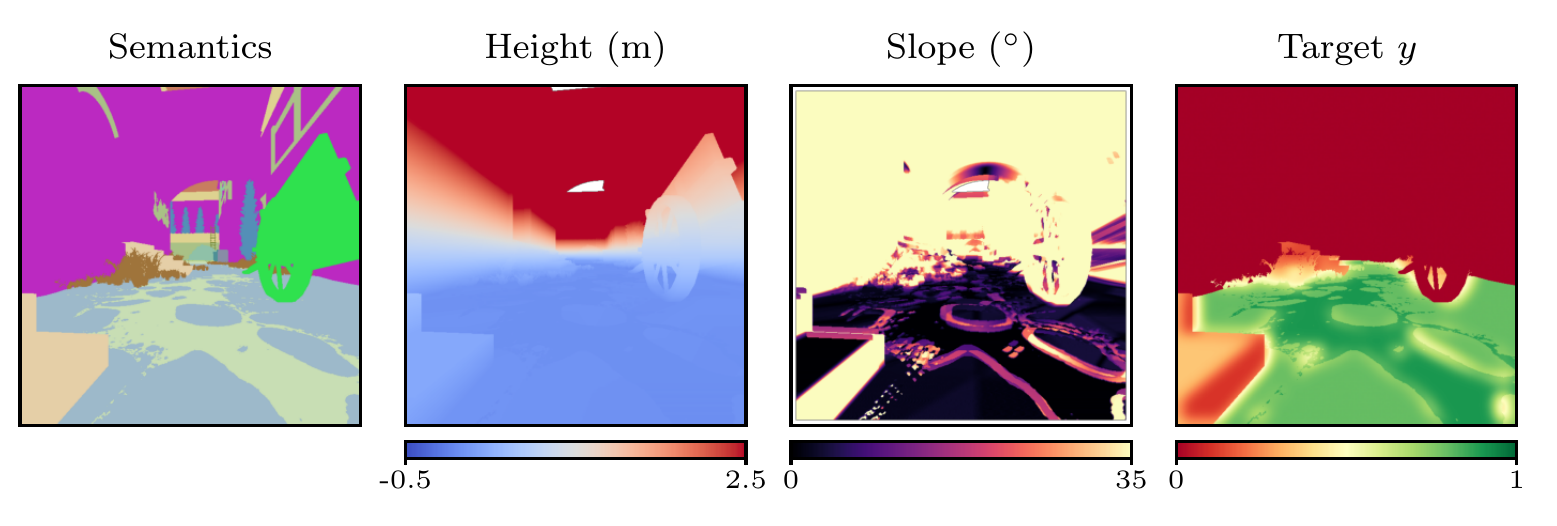}%
    \else\includegraphics[trim={176.6pt 0 0 0},clip,width=\linewidth]{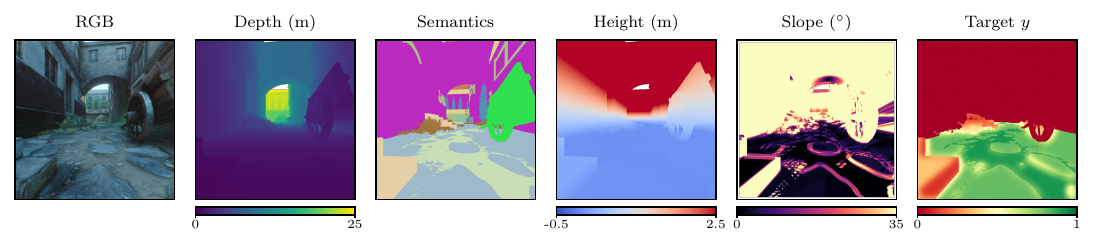}\fi
    \caption{Continuous traversability supervision on TartanGround. Left to right: the semantic labels for one frame, the geometry-derived height above ground and local surface slope, and the final target $y\in[0,1]$. The target combines a platform-grounded semantic prior with continuous geometric modulation, and cannot be trivially derived from semantics or geometry alone.}
    \label{fig:tartanground_img}
\end{figure}

\subsection{Dataset: TartanGround}

We train and evaluate on TartanGround \cite{patel2025tartangroundlargescaledatasetground}, a large-scale synthetic dataset spanning more than 20 photorealistic scene types, from construction sites and industrial facilities to forests, urban downtowns and underground mines. It provides temporally ordered RGB-D sequences from a front-facing camera on a simulated ground robot, with per-frame pixel-accurate semantic labels, random robot heights, and tracked camera pose.

\textbf{Split protocol.} We define a single canonical split at the trajectory level, shared by every model in this paper: of 136 trajectories across 24 environments, 94 are used for training, 20 for validation and model selection, and 22 are held out for testing. The split is stratified per environment, and two environments containing a single trajectory contribute only to training, so the test set spans the remaining 22. All reported results are computed exclusively on held-out test trajectories, which no model has seen during training or model selection.

\subsection{Continuous Traversability Labeling}
\label{subsec:labels}

Rather than constructing hard binary labels, we generate a continuous target $y\in[0,1]$ in four steps. We state the scope of this supervision at the outset: the targets encode semantic and geometric traversability, that is, whether a surface is of a drivable type and of a drivable shape, rather than a measurement of physical interaction. Self-supervised pipelines that derive labels from proprioceptive feedback observe whether a specific platform actually traversed a region, and so capture effects such as slip, sinkage and compliance that geometry and semantics cannot express; the cost is that they require deployment and only label terrain the robot has already driven over. The supervision here makes the opposite trade, covering every pixel of every frame offline while remaining an estimate of affordance rather than a record of it.

\textbf{Step 1: Camera height by per-trajectory self-calibration.} We estimate the camera height $h_c$ from the imagery itself, as the median vertical camera-frame coordinate over semantically-ground pixels in a near-field, bottom-centre region, aggregated across the trajectory. This makes the pipeline independent of pose annotation and platform metadata.

Fig.~\ref{fig:tartanground_img} shows the intermediate quantities of this process for a representative frame.

\textbf{Step 2: Height above ground.} TartanGround depth is planar $z$-depth from a pinhole camera, so the height above ground of pixel $(u,v)$ with depth $D_t(u,v)$ is
\begin{equation}
h_t(u,v) = h_c - \frac{(v - c_y)\, D_t(u,v)}{f_y},
\end{equation}
where $f_y$ and $c_y$ are the vertical focal length and principal-point offset of the camera, and $v$ is the image row. Since TartanGround randomises robot height per trajectory, exact projective geometry is what allows a fixed metric obstacle threshold to carry the same meaning throughout the dataset. Pixels with invalid depth retain the semantic prior alone.

\textbf{Step 3: Slope and roughness.} We unproject depth into a local 3D surface, estimate a unit normal $\mathbf{n}_t$ per valid pixel, and convert it to a slope angle
\begin{equation}
s_t(u,v)=\arccos\!\left(\left|\mathbf{n}_t(u,v)^\top\mathbf{g}\right|\right)\cdot\frac{180}{\pi},
\end{equation}
with $\mathbf{g}$ the gravity direction in camera coordinates. Normals are differenced over a wide stencil ($11$ px half-width) after median-filtering depth, so that the measurement reflects the grade of the terrain at the scale a robot traverses it. We also compute local roughness $r_t$ as the standard deviation of $h_t$ in a $9$ px window, separating a smooth ramp (high slope, low roughness) from rubble (moderate slope, high roughness).

\textbf{Step 4: Semantic prior and geometric modulation.} Each semantic class maps to a scalar prior through a single global lookup table $\mathrm{LUT}$ grounded in platform capability, $y_t^{\mathrm{sem}}=\mathrm{LUT}(S_t)$, where $S_t$ is the semantic label map of frame $t$. The table is specified once by hand, at the level of semantic classes rather than pixels, and is shared by every environment. Classes fall into capability tiers: engineered surfaces $1.00$, compacted ground $0.90$, firm natural $0.80$, vegetated $0.70$, loose debris $0.55$, dense low vegetation $0.40$, and blocked $0.00$. The values are ordinal, not calibrated: assigned by hand to span $[0,1]$ over the surfaces TartanGround contains, ordered by how a standard wheeled platform would cross them, and checked against the classes its recorded trajectories are actually driven over. The supervision relies on that ordering and on one shared table, not on the exact spacing.

Slope and roughness then attenuate this prior:
\begin{equation}
y_t = y_t^{\mathrm{sem}} \cdot \mathcal{G}\!\left[\rho(s_t)\,\rho(r_t)\right],
\end{equation}
where $\rho(\cdot)\in[\phi,1]$ falls linearly from $1$ to a floor $\phi$ as its argument passes from a free threshold to a limit. For slope $s_t$ these run from $12^\circ$ to $35^\circ$ with $\phi{=}0.3$, and for roughness $r_t$ from $0.03$ to $0.20$\,m with $\phi{=}0.4$. The operator $\mathcal{G}$ is a Gaussian blur of width $12$\,px applied to the modulation alone. Pixels where slope or roughness is unmeasurable take $\rho=1$, so a missing cue never penalises the prior.

The floors keep geometry refining the prior rather than overriding it, which matters on vegetation, where per-pixel normals read steep over drivable ground; smoothing only the modulation leaves class boundaries sharp.

Finally, $y_t=0$ where terrain is unreachable: where a pixel stands more than $0.55$\,m above the local terrain trend $\overline{h_t}$, a $65$\,px box mean of $h_t$, which isolates obstacles independently of overall grade; or beyond a permissive global band ($h_t>2.5$\,m, or $h_t<-1.0$\,m for negative obstacles) that rejects clearly elevated structure such as overpasses.

Unlike the unordered images of FroboBots-2K~\cite{wang2025geniegeneralizablenavigationinthewild}, TartanGround supplies the contiguous sequences SAM2's video memory requires, and the graded target lets a deployed platform choose its own operating point within the clearance envelope Step 4's constraints impose.

\subsection{Training Details}

All models are trained on TartanGround using sequences of $T{=}24$ frames sampled at a stride of 2 (approximately 5 seconds of robot motion at 10 fps). For SAM-VTP, the first $N_\mathrm{warmup}$ frames populate the memory bank and are excluded from the loss. We set $N_\mathrm{warmup}{=}2$, since the bank is addressable once non-empty and requiring it full would never train the partially-populated regime deployment begins in. We use AdamW (base $\mathrm{lr}{=}3{\times}10^{-5}$, weight decay $10^{-3}$) with cosine annealing and warm restarts (period 15 epochs), automatic mixed precision (bfloat16) and gradient accumulation, on an RTX 5090 GPU. Augmentation parameters are sampled once per clip so that every frame of a clip receives the same transform, and checkpoints are selected by a rolling average of validation fuzzy Jaccard.

For deployment, both SAM-TP and SAM-VTP are made CUDA Graph compatible, which SAM2 does not natively support \cite{ravi2024sam2}. Replaying the captured graph amortises kernel-launch overhead at every timestep.

\subsection{Baselines}

\subsubsection{Re-implementation of SAM-TP}
\label{subsec:optimisation}

Comparing latency against the publicly released GeNIE implementation would measure implementation overhead rather than architecture. That inference path routes images through \texttt{SAM2ImagePredictor}, which applies its own normalisation pipeline and adds Python overhead, and leaves SAM2's memory components active though no temporal context is used. We therefore re-implement the baseline to call the SAM2 backbone and decoder directly with the memory pathway disabled, which reduces per-frame latency from $21.9$\,ms to $2.6$\,ms in eager execution, and to $1.9$\,ms ($540$\,fps) under CUDA Graph replay. This is not a contribution of this work; it is what makes the reported frame-wise cost attributable to frame-wise computation itself.

\subsubsection{Fair Comparison via TartanGround Fine-Tuning}

Each baseline is fine-tuned on TartanGround's continuous traversability maps before evaluation, since published weights applied out-of-the-box degrade substantially under the visual domain gap and the differing label ontology. Every trainable model sees the same split and label version and is selected by the same rule, so observed differences reflect architecture rather than supervision or checkpoint choice.

\textbf{GeNIE (SAM-TP)}\cite{wang2025geniegeneralizablenavigationinthewild} is fine-tuned here as a supervised frame-wise SAM2 decoder baseline (no temporal memory). \textbf{WVN}\cite{mattamala2024wildvisualnavigationfast} and \textbf{WayFASTER} \cite{gasparino2024wayfasterselfsupervisedtraversabilityprediction} were designed for self-supervised traversability learning; for a controlled comparison we keep their architectures but replace those objectives with direct supervision from the same continuous masks. WVN then remains frame-wise and WayFASTER retains ConvLSTM aggregation, but neither maintains a persistent, addressable cross-frame memory bank. We additionally report GeNIE's released SAM-TP weights zero-shot, without TartanGround fine-tuning, as a cross-dataset transfer reference.

\subsection{Metrics}

Our primary accuracy metric is the \textbf{fuzzy Jaccard} similarity $J(p,g)$ of Eq.~\eqref{eq:ruzicka}, which equals $1$ if and only if the prediction matches the target at every pixel and is therefore proper for continuous supervision.

\textbf{MAE and signed bias.} Mean absolute error over the $M$ evaluated pixels measures average per-pixel deviation; because it is symmetric we also report the signed bias $\frac{1}{M}\sum_i (p_i - g_i)$, separating a model that over-predicts traversability from one that is merely imprecise in both directions.

\textbf{Calibration.} We adapt Expected Calibration Error to soft targets, binning pixels into $K{=}15$ equal-width bins on predicted probability and taking the $|B_k|/M$-weighted absolute difference between the mean prediction and the mean target within each bin. A low ECE means a predicted score of $0.6$ denotes terrain of traversability $0.6$, which is what makes a continuous output usable for platform-specific thresholding.

\textbf{Thresholded IoU.} We retain thresholded (binary) IoU as a deployment-oriented diagnostic. For each model the threshold is selected independently, by sweeping $[0.1,0.9]$ in steps of $0.05$ on the validation trajectories and retaining the value that maximises IoU there; it is then held fixed for evaluation on the test trajectories. Selecting it per model avoids penalising differences in predicted score scale, while fixing it before evaluation keeps the reported figure free of test-set tuning.

We do not report Soft-IoU: it carries the same over-prediction bias as the region losses rejected in Eq.~\eqref{eq:ruzicka}.

\section{Results}
\label{sec:results}

We evaluate SAM-VTP against SAM-TP, WVN, and WayFASTER across accuracy, temporal consistency, component ablations, closed-loop navigation, and edge-device deployment. The SAM-TP and SAM-VTP rows of Table~\ref{tab:tartanground_overall} are means over three independently trained models; seed-to-seed standard deviation stays below $0.001$ on every metric (fJacc $0.9077 \pm 0.0003$ against $0.9241 \pm 0.0009$), placing the accuracy gap at over twenty times the run-to-run spread and the two temporal gaps at seventeen and nineteen times it. The other rows are single runs.

\begin{table*}[!htbp]
\setcounter{table}{0}
\centering
\caption{Accuracy, temporal consistency, and throughput on held-out TartanGround test trajectories. All models fine-tuned on TartanGround unless noted. Best in \textbf{bold}. fJacc is the fuzzy (Ruzicka) Jaccard index over continuous scores and is the primary metric; IoU is the binary Jaccard index at a per-model threshold calibrated on validation; ECE is the expected calibration error against the continuous targets. Stab.\ and Flick.\ are the mean per-sequence standard deviation of fJacc across frames and its mean absolute frame-to-frame change, both lower-is-better; basing them on fJacc rather than a thresholded score keeps them on the primary metric. Throughput uses each model's own deployment resolution ($512{\times}512$ except WayFASTER, which predicts a $100{\times}100$ BEV grid from $320{\times}180$ input; with targets interpolated to that grid and predictions upsampled back, limiting its attainable fidelity).}
\label{tab:tartanground_overall}
\iflatexml\else\small\fi
\iflatexml
\setlength{\tabcolsep}{5pt}
\begin{tabular}{lccccccccc}
\toprule
& \multicolumn{4}{c}{\textbf{Accuracy}} & \multicolumn{2}{c}{\textbf{Temporal}} & \multicolumn{3}{c}{\textbf{Throughput (FPS)}$\uparrow$} \\
\cmidrule(lr){2-5}\cmidrule(lr){6-7}\cmidrule(lr){8-10}
\textbf{Model} & \textbf{fJacc}$\uparrow$ & \textbf{MAE}$\downarrow$ & \textbf{ECE}$\downarrow$ & \textbf{IoU}$\uparrow$
& \textbf{Stab.}$\downarrow$ & \textbf{Flick.}$\downarrow$
& \begin{tabular}{@{}c@{}}\textbf{RTX}\\\textbf{5090}\end{tabular}
& \begin{tabular}{@{}c@{}}\textbf{AGX}\\\textbf{Orin}\end{tabular}
& \begin{tabular}{@{}c@{}}\textbf{Orin}\\\textbf{Nano}\end{tabular} \\
\midrule
WayFASTER      & 0.791 & 0.066 & 0.039 & 0.740 & 0.0630 & 0.0264 & 456 & \textbf{52.8} & 26.8 \\
WVN            & 0.760 & 0.079 & 0.048 & 0.741 & 0.0597 & 0.0246 & 19.5 & --$^\ddagger$ & --$^\ddagger$ \\
SAM-TP         & 0.908 & 0.024 & 0.013 & 0.854 & 0.0387 & 0.0175 & \textbf{540*} & 44.9* & \textbf{33.5*} \\
SAM-VTP~(ours) & \textbf{0.924} & \textbf{0.018} & \textbf{0.010} & \textbf{0.886} & \textbf{0.0323} & \textbf{0.0120} & 296 & 30.1 & 21.4 \\
\midrule
SAM-TP~(GeNIE)$^\dagger$ & 0.667 & 0.126 & 0.111 & 0.678 & 0.0908 & 0.0452 & 45.7 & 6.3 & 3.6 \\
\bottomrule
\end{tabular}

*Optimised SAM-TP implementation, CUDA-graph replayed; see Sec.~\ref{subsec:optimisation}. Baselines are timed as released and run eager, which is why GeNIE is $7\times$ slower than SAM-TP despite sharing its architecture.
$^\dagger$Original GeNIE weights, trained on binary Frodobots-2K labels.
$^\ddagger$Not deployed to the embedded devices: its STEGO extractor is already the slowest model here.
\else
\setlength{\tabcolsep}{5pt}
\sbox0{\begin{tabular}{lccccccccc}
\toprule
& \multicolumn{4}{c}{\textbf{Accuracy}} & \multicolumn{2}{c}{\textbf{Temporal}} & \multicolumn{3}{c}{\textbf{Throughput (FPS)}$\uparrow$} \\
\cmidrule(lr){2-5}\cmidrule(lr){6-7}\cmidrule(lr){8-10}
\textbf{Model} & \textbf{fJacc}$\uparrow$ & \textbf{MAE}$\downarrow$ & \textbf{ECE}$\downarrow$ & \textbf{IoU}$\uparrow$
& \textbf{Stab.}$\downarrow$ & \textbf{Flick.}$\downarrow$ & \textbf{RTX 5090} & \textbf{AGX Orin} & \textbf{Orin Nano} \\
\midrule
WayFASTER      & 0.791 & 0.066 & 0.039 & 0.740 & 0.0630 & 0.0264 & 456 & \textbf{52.8} & 26.8 \\
WVN            & 0.760 & 0.079 & 0.048 & 0.741 & 0.0597 & 0.0246 & 19.5 & --$^\ddagger$ & --$^\ddagger$ \\
SAM-TP         & 0.908 & 0.024 & 0.013 & 0.854 & 0.0387 & 0.0175 & \textbf{540*} & 44.9* & \textbf{33.5*} \\
SAM-VTP (ours) & \textbf{0.924} & \textbf{0.018} & \textbf{0.010} & \textbf{0.886} & \textbf{0.0323} & \textbf{0.0120} & 296 & 30.1 & 21.4 \\
\midrule
SAM-TP (GeNIE)$^\dagger$ & 0.667 & 0.126 & 0.111 & 0.678 & 0.0908 & 0.0452 & 45.7 & 6.3 & 3.6 \\
\bottomrule
\end{tabular}}%
\usebox0
\par\vspace*{3pt}
\parbox{\wd0}{\footnotesize
*Optimised SAM-TP implementation, CUDA-graph replayed; see Sec.~\ref{subsec:optimisation}. Baselines are timed as released and run eager, which is why GeNIE is $7\times$ slower than SAM-TP despite sharing its architecture.
$^\dagger$Original GeNIE weights, trained on binary Frodobots-2K labels.
$^\ddagger$Not deployed to the embedded devices: its STEGO extractor is already the slowest model here.}
\fi
\end{table*}

SAM-VTP leads every accuracy and temporal measure in Table~\ref{tab:tartanground_overall}. It improves fuzzy Jaccard by 0.016 over the frame-wise SAM-TP baseline, which receives RGB alone, against a seed standard deviation below 0.001 on both arms (SAM-TP and SAM-VTP), and cuts expected calibration error by a fifth. Both SAM arms sit well clear of WayFASTER and WVN on every measure. GeNIE's released weights transfer least well of all and are the worst calibrated, with an optimal threshold of 0.15: their deficit is one of supervision and domain, not capacity, since the architecture is SAM-TP's.

The temporal margin holds in-domain: flicker falls by 31\% and per-sequence stability by 17\%, against a seed spread under $0.0004$ on both, and widens on the robot, where the same pair separates by 4.3$\times$ on flicker over the predicted region (Table~\ref{tab:closed_loop}). Much of what remains is scene-driven rather than model-driven: scored on identical frames the two arms' frame-to-frame changes correlate at $r{=}0.83$, and memory removes 65\% of the model-attributable part.

\begin{table}[!htbp]
\setcounter{table}{1}
\centering
\caption{Ablation on held-out TartanGround test trajectories, scored over 1000 sequence windows (24k frames). The first four rows form a $2\times2$ factorial over the two proposed components; SAM-TP\,+\,geometry is the fairness control, giving the frame-wise baseline the same depth-derived input as SAM-VTP. The last two rows withhold depth at inference and replace continuous targets with binary ones. Training budgets here are shorter than in Table~\ref{tab:tartanground_overall}, so absolute scores are lower and the two tables are not directly comparable. Best in \textbf{bold}.}
\label{tab:ablation}
\iflatexml\setlength{\tabcolsep}{5pt}\else\footnotesize
\setlength{\tabcolsep}{2.5pt}\fi
\begin{tabular}{@{}lcccccc@{}}
\toprule
& \multicolumn{2}{c}{\textbf{Component}} & & & & \\
\cmidrule(lr){2-3}
\textbf{Configuration} & \textbf{Mem.} & \textbf{Geo.} & \textbf{fJacc}$\uparrow$ & \textbf{MAE}$\downarrow$ & \textbf{ECE}$\downarrow$ & \textbf{Flick.}$\downarrow$ \\
\midrule
SAM-TP (frame-wise)   & --         & --         & 0.875 & 0.033 & 0.019 & 0.022 \\
\quad + geometry      & --         & \checkmark & 0.900 & 0.026 & 0.014 & 0.019 \\
SAM-VTP (memory)      & \checkmark & --         & 0.894 & 0.028 & 0.016 & 0.015 \\
\quad + geometry (full) & \checkmark & \checkmark & \textbf{0.914} & \textbf{0.022} & \textbf{0.013} & \textbf{0.013} \\
\midrule
Full, depth withheld  & \checkmark & --         & 0.853 & 0.042 & 0.031 & 0.020 \\
Full, binary targets  & \checkmark & \checkmark & 0.733 & 0.072 & 0.069 & 0.029 \\
\bottomrule
\end{tabular}
\end{table}

Table~\ref{tab:ablation} attributes SAM-VTP's advantage to its two components. Differences are paired at the level of the sequence window rather than the frame, since the 24 frames of a window are consecutive views of one scene and treating them as independent inflates significance roughly threefold; at 1000 windows the minimum detectable difference in fJacc is $0.005$, so every contrast below is resolved with margin.

Table~\ref{tab:ablation} trains on a shorter budget, where the same contrast is larger ($0.875\rightarrow0.914$ against $0.908\rightarrow0.924$): part of what the components buy is convergence speed, so the decomposition below holds at this budget rather than the fully-trained one. Both components contribute and neither subsumes the other. Geometry alone raises fJacc by $0.025$ and memory alone by $0.019$, and each still pays on top of the other: memory adds $0.014$ to the geometry-equipped baseline, geometry adds $0.020$ to the memory-only model.

\begin{figure}[!t]
\centering
\iflatexml\includegraphics[width=0.75\linewidth]{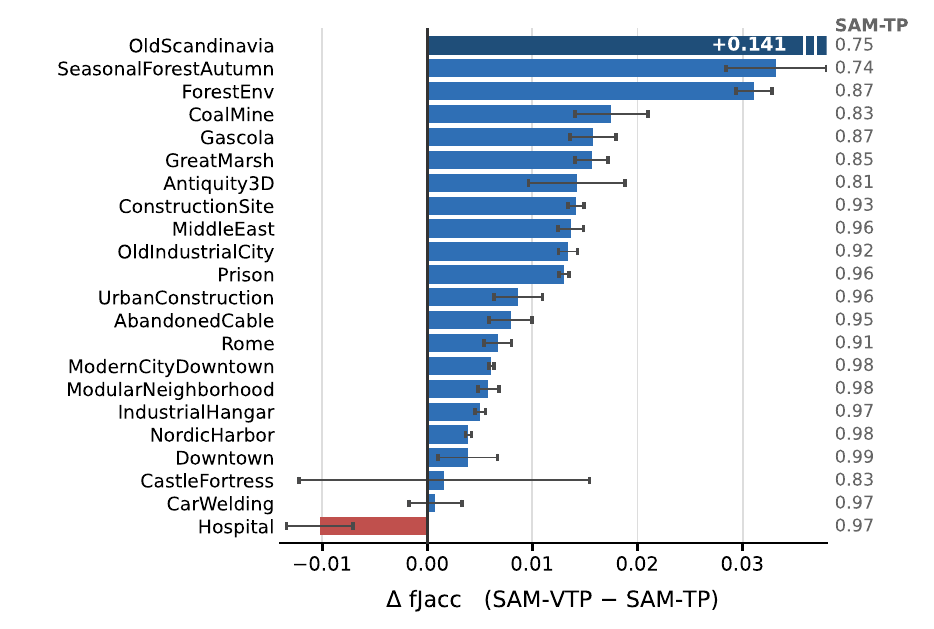}%
\else\includegraphics[width=0.86\linewidth]{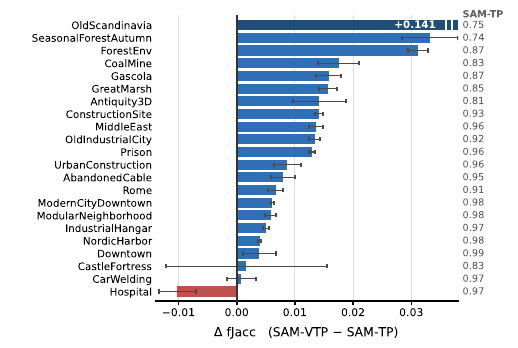}\fi
\caption{Per-environment difference in fJacc between SAM-VTP and SAM-TP on the 22 held-out test trajectories, sorted by magnitude; bars are the mean over three seeds and error bars their standard deviation. SAM-VTP is ahead in 21 of 22 environments. The grey column gives SAM-TP's own fJacc, showing that the largest gains fall where the frame-wise baseline is weakest ($r=-0.61$) while environments it already scores above $0.95$ gain least. OldScandinavia is clipped at the axis limit; its value is $+0.141$.}
\label{fig:per_env_delta}
\end{figure}

The components also differ in cost. On the Orin Nano, memory adds $14.9$\,ms per frame and geometry $2.1$\,ms, so geometry yields roughly ten times more accuracy per millisecond. Memory's cost is justified by the hardest frames rather than the average: ranking frames by difficulty, the hardest tenth account for 38\% of its gain and the easiest tenth for 1\%, the proportion of frames below $0.8$ fJacc falls by a fifth, and first-percentile accuracy more than doubles. Fig.~\ref{fig:per_env_delta} shows the same concentration across environments: the largest gains fall in unstructured, vegetated terrain such as forest, marsh and mine, and the smallest in structured built environments where the frame-wise baseline is already near ceiling.

Memory is the larger contributor to temporal steadiness, but not the only one: frame-to-frame change falls from $0.022$ to $0.015$ once memory is enabled, and geometry adds a smaller reduction in both the frame-wise ($0.022\rightarrow0.019$) and memory-equipped ($0.015\rightarrow0.013$) model. Withholding depth raises flicker to $0.020$, so steadiness depends on it too.

Flicker alone, however, does not justify memory. A causal exponential moving average (EMA) over the frame-wise model's outputs reaches lower flicker than SAM-VTP at every smoothing factor we tested, but only at a substantial cost in accuracy. Averaging pixel-aligned probabilities blurs boundaries across a moving camera, whereas memory attention associates by content. We therefore report flicker only alongside accuracy, since in isolation it is optimised by making the model worse.

Continuous supervision is not cosmetic, and the metric it costs is instructive. Training the identical architecture on binarised targets improves thresholded binary IoU by $0.012$, from $0.875$ to $0.887$ at each model's own tuned threshold, which is unsurprising since that is the objective it optimises, while fJacc falls by $0.181$, MAE more than triples, and ECE rises from $0.013$ to $0.069$. A binary-trained model is therefore competitive only at one operating point and poorly calibrated elsewhere, a trade invisible to any fixed-threshold evaluation and not worth making for a planner consuming graded cost.

\begin{table}[!htbp]
\setcounter{table}{2}
\centering
\caption{Closed-loop navigation on a physical robot at two field sites: a paved course with mixed surfaces and vegetation (15 runs), and rough, undulating, overgrown ground (21 runs). The baseline is SAM-TP trained on the same data, objective and splits, with planner and controller shared and trials interleaved between models at both sites. SAM-TP takes RGB only and SAM-VTP takes RGB-D, so the table compares the two systems and does not separate memory from depth. Flicker is the mean absolute frame-to-frame change in predicted probability over the predicted region, the union of $P>0.5$ across the pair, as are traversable fraction and null predictions; both arms select $P>0.5$ independently on validation, and the ordering holds for any threshold in $[0.1,0.8]$. Every row separates completely on the course; on rough ground no-path, traversable fraction and flicker separate in all ten paired trials, a sign test at its floor ($p=0.002$), while progress rate and distance overlap ($8/10$ and $7/10$ pairs).}
\label{tab:closed_loop}
\iflatexml\setlength{\tabcolsep}{5pt}\else\footnotesize
\setlength{\tabcolsep}{3pt}\fi
\begin{tabular}{@{}lccccccc@{}}
\toprule
& \multicolumn{3}{c}{\textbf{Paved course}} & \multicolumn{3}{c}{\textbf{Rough terrain}} \\
\cmidrule(lr){2-4}\cmidrule(lr){5-7}
\textbf{Measure} & TP & VTP & Ratio & TP & VTP & Ratio \\
\midrule
No-path (\%)$\downarrow$          & 32.4 & \textbf{9.9}   & 0.31$\times$ & 20.6 & \textbf{3.0}   & 0.15$\times$ \\
Traversable frac.$\uparrow$       & 0.225 & \textbf{0.338} & 1.50$\times$ & 0.076 & \textbf{0.216} & 2.86$\times$ \\
Flicker (region)$\downarrow$      & 0.129 & \textbf{0.030} & 0.23$\times$ & 0.055 & \textbf{0.036} & 0.64$\times$ \\
Null predictions (\%)$\downarrow$ & 7.9 & \textbf{0.9}     & 0.12$\times$ & 4.2 & \textbf{0.8}     & 0.18$\times$ \\
Progress rate (m/s)$\uparrow$    & 0.141 & \textbf{0.182} & 1.29$\times$ & 0.159 & \textbf{0.180} & 1.13$\times$ \\
Distance (m)$\uparrow$            & 7.34 & \textbf{15.84}  & 2.16$\times$ & 14.59 & \textbf{18.68} & 1.28$\times$ \\
\bottomrule
\end{tabular}
\end{table}

\begin{figure}[!t]
    \centering
    \iflatexml\includegraphics[width=\linewidth]{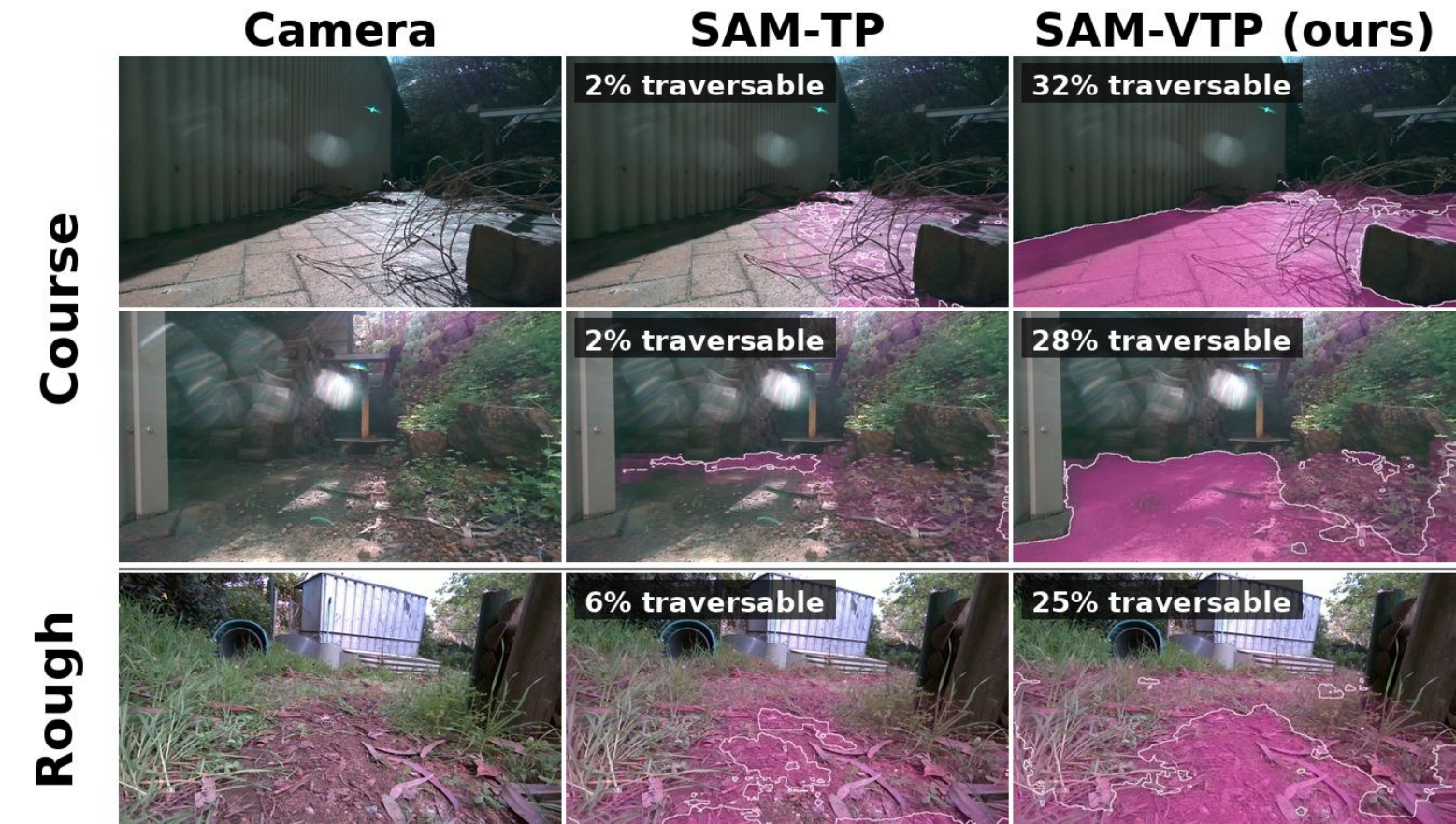}%
    \else\includegraphics[width=0.90\linewidth]{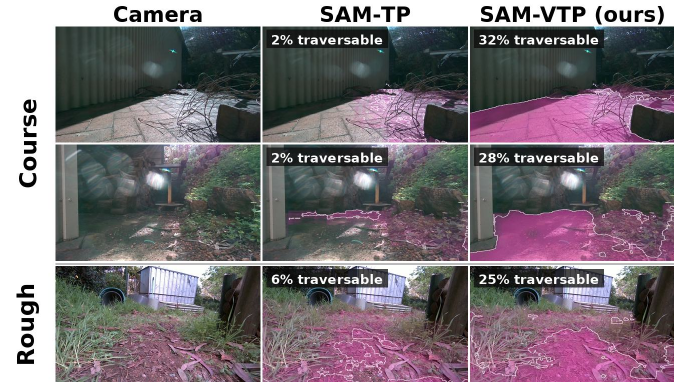}\fi
    \caption{Both models replayed over the same recorded frames in the deployed configuration (SAM-TP on RGB, SAM-VTP on RGB-D), so both see the same scene. Overlay opacity is the predicted probability; the white contour marks $P>0.5$. The top two rows are the paved course, where SAM-TP breaks into disconnected fragments around clutter (2\% against 32\%) and again at the terrain and lighting transition at which its trials ended, where the planner was reporting no feasible path (2\% against 28\%); SAM-VTP holds a single coherent region in both. The bottom row is rough terrain, where SAM-TP under-segments a clearly defined path (6\% against 25\%) rather than fragmenting.}
    \label{fig:qual_field}
\end{figure}

The platform is an Ackermann-steered JetAcker with an Intel RealSense D455f, running all perception onboard a Jetson Orin Nano 8\,GB (JetPack 6.2). Both arms drive GeNIE's arc planner and a pure-pursuit controller. A safety gate halts the platform when the planner returns no feasible path or perception goes stale, so lost traversable area becomes lost motion. Trials alternate between models: the rough site yields ten complete pairs, the sign test's $n$, while the course's fifteen runs do not pair evenly. An accompanying video shows interleaved trials at both sites. Table~\ref{tab:closed_loop} reports two sites: a paved course containing visually ambiguous obstacles, changes in surface and lighting, and partially transparent vegetation that must be driven around rather than through; and rough, undulating, overgrown ground away from any hand-laid course. The arc planner finds no feasible path on 32.4\% of cycles against 9.9\%, and the safety gate blocks the platform for correspondingly longer. Each arm drives its own trajectory, so the perception rows are not scored on identical input; replaying both models over three trials per site, from both arms' paths, preserves every ordering (coverage $1.27\times$ and $2.12\times$, flicker $0.22\times$ and $0.60\times$), and Fig.~\ref{fig:qual_field} shows one such replay.

Runs were ended by the operator, not at a fixed goal: SAM-TP was typically retrieved after the safety gate had blocked the platform, whereas SAM-VTP was still driving when stopped. Progress rate is the distance travelled divided by the driving window, measured from release to the end of the run with blocked time included, so time spent halted lowers the rate rather than being removed from it. Both separate on the course, where SAM-TP's runs stopped at the transition of Fig.~\ref{fig:qual_field} rather than degrading gradually; neither does on rough ground ($8/10$ and $7/10$ pairs), where the chassis high-centres on grass tufts and clearance, not perception, becomes binding. Both are integrated from commanded speed, since the chassis exposes no wheel encoders, so wheel spin while high-centred on the rough site counts as travel.

The throughput columns describe the deployed system rather than a benchmark. The Orin Nano is the module the robot carries, so the closed-loop trials ran at the $21.4$ FPS Table~\ref{tab:tartanground_overall} reports there. Both SAM arms clear real time on both embedded devices; GeNIE's released implementation reaches $3.6$ FPS on the Nano, which is why the frame-wise baseline is re-implemented (Section~\ref{subsec:optimisation}).

\section{Conclusion}
\label{sec:conclusion}

This paper presents SAM-VTP, a memory-augmented traversability segmentation model that leverages SAM2 video memory to condition each prediction on recent observations. The results show that modelling traversability over a short temporal window improves performance relative to independent frame-wise prediction.

On TartanGround, SAM-VTP outperforms the frame-wise SAM-TP baseline across all accuracy and temporal consistency metrics, and the gain carries into navigation: driving the same planner and controller, a physical robot stalls for want of a feasible path a third as often on a paved course and a seventh as often on rough ground. The model runs at 30 FPS on a NVIDIA Jetson AGX Orin, and at $21.4$ FPS on the Orin Nano the robot carries.

At the ablation's training budget, geometry gives the larger accuracy gain and memory the larger gain in temporal consistency. Each still adds to the other, though together they fall slightly short of the sum of their separate gains ($0.039$ fJacc against $0.044$).

Limitations remain. The deployed system requires a depth sensor, as withholding depth at inference costs $0.061$ fJacc. The targets commit to a clearance envelope, so a platform differing substantially in clearance requires them to be regenerated rather than merely re-thresholded, and sequence initialisation introduces a brief warm-up period. All segmentation accuracy reported here is synthetic, scored against our own pipeline's targets; the closed-loop trials measure planner behaviour, not mask quality. The pipeline also requires a per-pixel semantic map, which the simulator supplies and real footage does not, so applying it outside simulation needs a semantic segmenter in the loop. Future work will focus on expanding real-world data, uncertainty calibration, and reducing cold-start latency.

\bibliographystyle{IEEEtran}
\bibliography{main}

\begin{thebibliography}{10}
\providecommand{\url}[1]{#1}
\csname url@samestyle\endcsname
\providecommand{\newblock}{\relax}
\providecommand{\bibinfo}[2]{#2}
\providecommand{\BIBentrySTDinterwordspacing}{\spaceskip=0pt\relax}
\providecommand{\BIBentryALTinterwordstretchfactor}{4}
\providecommand{\BIBentryALTinterwordspacing}{\spaceskip=\fontdimen2\font plus
\BIBentryALTinterwordstretchfactor\fontdimen3\font minus
  \fontdimen4\font\relax}
\providecommand{\BIBforeignlanguage}[2]{{%
\expandafter\ifx\csname l@#1\endcsname\relax
\typeout{** WARNING: IEEEtran.bst: No hyphenation pattern has been}%
\typeout{** loaded for the language `#1'. Using the pattern for}%
\typeout{** the default language instead.}%
\else
\language=\csname l@#1\endcsname
\fi
#2}}
\providecommand{\BIBdecl}{\relax}
\BIBdecl

\bibitem{wang2025geniegeneralizablenavigationinthewild}
J.~Wang \emph{et~al.}, ``Genie: A generalizable navigation system for
  in-the-wild environments,'' \emph{IEEE Robotics and Automation Letters},
  2025.

\bibitem{zhang2025dualbevnavduallayerbevbased}
J.~Zhang \emph{et~al.}, ``Dual-bev nav: Dual-layer bev-based heuristic path
  planning for robotic navigation in unstructured outdoor environments,'' in
  \emph{IEEE International Conference on Robotics and Automation (ICRA)}, 2025,
  pp. 8872--8879.

\bibitem{frey2024roadrunnerlearningtraversability}
J.~Frey \emph{et~al.}, ``Roadrunner -- learning traversability estimation for
  autonomous off-road driving,'' \emph{IEEE Transactions on Field Robotics},
  2024.

\bibitem{mattamala2024wildvisualnavigationfast}
M.~Mattamala \emph{et~al.}, ``Wild visual navigation: Fast traversability
  learning via pre-trained models and online self-supervision,''
  \emph{Autonomous Robots}, 2025.

\bibitem{annurev:/content/journals/10.1146/annurev-control-062722-100728}
T.~H. Chung, V.~Orekhov, and A.~Maio, ``Into the robotic depths: Analysis and
  insights from the darpa subterranean challenge,'' \emph{Annual Review of
  Control, Robotics, and Autonomous Systems}, vol.~6, no. Volume 6, 2023, pp.
  477--502, 2023.

\bibitem{sevastopoulos2022surveytraversabilityestimationmobile}
C.~Sevastopoulos and S.~Konstantopoulos, ``A survey of traversability
  estimation for mobile robots,'' \emph{IEEE Access}, 2022.

\bibitem{kirillov2023segment}
A.~Kirillov \emph{et~al.}, ``Segment anything,'' in \emph{Proceedings of the
  IEEE/CVF International Conference on Computer Vision (ICCV)}, 2023.

\bibitem{ravi2024sam2}
N.~Ravi \emph{et~al.}, ``Sam 2: Segment anything in images and videos,'' in
  \emph{International Conference on Learning Representations (ICLR)}, 2025.

\bibitem{palazzo2020domainadaptationoutdoorrobot}
S.~Palazzo \emph{et~al.}, ``Domain adaptation for outdoor robot traversability
  estimation from rgb data with safety-preserving loss,'' in \emph{IEEE/RSJ
  International Conference on Intelligent Robots and Systems (IROS)}, 2020.

\bibitem{Vecchio2024}
G.~Vecchio, S.~Palazzo, D.~C. Guastella, D.~Giordano, G.~Muscato, and
  C.~Spampinato, ``Terrain traversability prediction through self-supervised
  learning and unsupervised domain adaptation on synthetic data,''
  \emph{Autonomous Robots}, vol.~48, no.~2, p.~4, 2024.

\bibitem{jung2024vstrongvisualselfsupervisedtraversability}
S.~Jung, J.~Lee, X.~Meng, B.~Boots, and A.~Lambert, ``V-strong: Visual
  self-supervised traversability learning for off-road navigation,'' in
  \emph{IEEE International Conference on Robotics and Automation (ICRA)}, 2024.

\bibitem{sahu2025anytraverseoffroadtraversabilityframework}
S.~Sahu, A.~Singh, K.~Nambiar, S.~Saripalli, and P.~B. Sujit, ``Anytraverse: An
  off-road traversability framework with vlm and human operator in the loop,''
  2025, arXiv:2506.16826.

\bibitem{patel2024roadrunnermmlearning}
M.~Patel, J.~Frey, D.~Atha, P.~Spieler, M.~Hutter, and S.~Khattak, ``Roadrunner
  m\&m -- learning multi-range multi-resolution traversability maps for
  autonomous off-road navigation,'' \emph{IEEE Robotics and Automation
  Letters}, 2024.

\bibitem{BaghbaderaniLW024}
R.~K. Baghbaderani, Y.~Li, S.~Wang, and H.~Qi, ``Temporally-consistent video
  semantic segmentation with bidirectional occlusion-guided feature
  propagation,'' in \emph{WACV}, 2024, pp. 674--684.

\bibitem{shaban2021semantic}
A.~Shaban, X.~Meng, J.~Lee, B.~Boots, and D.~Fox, ``Semantic terrain
  classification for off-road autonomous driving,'' in \emph{5th Annual
  Conference on Robot Learning}, 2021.

\bibitem{Gasparino_2022}
M.~V. Gasparino \emph{et~al.}, ``Wayfast: Navigation with predictive
  traversability in the field,'' \emph{IEEE Robotics and Automation Letters},
  vol.~7, no.~4, p. 10651–10658, Oct. 2022.

\bibitem{gasparino2024wayfasterselfsupervisedtraversabilityprediction}
M.~V. Gasparino, A.~N. Sivakumar, and G.~Chowdhary, ``Wayfaster: a
  self-supervised traversability prediction for increased navigation
  awareness,'' in \emph{IEEE International Conference on Robotics and
  Automation (ICRA)}, 2024.

\bibitem{guan2022tnsterraintraversabilitymapping}
T.~Guan, Z.~He, R.~Song, D.~Manocha, and L.~Zhang, ``Tns: Terrain
  traversability mapping and navigation system for autonomous excavators,'' in
  \emph{Robotics: Science and Systems (RSS)}, 2022.

\bibitem{zhang2024traversabilityawareleggednavigationlearning}
H.~Zhang \emph{et~al.}, ``Traversability-aware legged navigation by learning
  from real-world visual data,'' \emph{IEEE Transactions on Robotics}, 2026.

\bibitem{s23063239}
M.~Sánchez, J.~Morales, and J.~L. Martínez, ``Reinforcement and curriculum
  learning for off-road navigation of an ugv with a 3d lidar,'' \emph{Sensors},
  vol.~23, no.~6, 2023.

\bibitem{chen2023fastc}
Y.~Chen, P.~Wei, Z.~Liu, B.~Wang, J.~Yang, and W.~Liu, ``{FASTC}: A fast
  attentional framework for semantic traversability classification using point
  cloud,'' in \emph{Proceedings of the 26th European Conference on Artificial
  Intelligence (ECAI)}, ser. Frontiers in Artificial Intelligence and
  Applications.\hskip 1em plus 0.5em minus 0.4em\relax IOS Press, 2023.

\bibitem{capezzuto2026cat}
G.~Capezzuto, F.~Tommaselli, M.~P. Angarola, R.~V. Godoy, and M.~Becker,
  ``Towards capability-aware traversability navigation for unstructured
  environments,'' in \emph{IEEE/RSJ International Conference on Intelligent
  Robots and Systems (IROS)}, 2026.

\bibitem{Giusti2016AML}
A.~Giusti \emph{et~al.}, ``A machine learning approach to visual perception of
  forest trails for mobile robots,'' \emph{IEEE Robotics and Automation
  Letters}, vol.~1, pp. 661--667, 2016.

\bibitem{jeon2024followfootprintsselfsupervisedtraversability}
Y.~Jeon, E.~I. Son, and S.-W. Seo, ``Follow the footprints: Self-supervised
  traversability estimation for off-road vehicle navigation based on geometric
  and visual cues,'' in \emph{IEEE International Conference on Robotics and
  Automation (ICRA)}, 2024.

\bibitem{castro2023doesfeelselfsupervisedcostmap}
M.~G. Castro \emph{et~al.}, ``How does it feel? self-supervised costmap
  learning for off-road vehicle traversability,'' in \emph{IEEE International
  Conference on Robotics and Automation (ICRA)}, 2023.

\bibitem{sivaprakasam2024salonselfsupervisedadaptivelearning}
M.~Sivaprakasam \emph{et~al.}, ``Salon: Self-supervised adaptive learning for
  off-road navigation,'' in \emph{IEEE International Conference on Robotics and
  Automation (ICRA)}, 2025.

\bibitem{Rafi2024}
T.~H. Rafi, R.~Mahjabin, E.~Ghosh, Y.-W. Ko, and J.-G. Lee, ``Domain
  generalization for semantic segmentation: a survey,'' \emph{Artificial
  Intelligence Review}, vol.~57, no.~9, p. 247, 2024.

\bibitem{chen2022calicoarsetofinealignmentsbased}
Z.~Chen, D.~Pushp, and L.~Liu, ``Cali: Coarse-to-fine alignments based
  unsupervised domain adaptation of traversability prediction for deployable
  autonomous navigation,'' in \emph{Robotics: Science and Systems (RSS)}, 2022.

\bibitem{hwang2026vita}
J.-H. Hwang, J.~Bae, D.-W. Kim, Y.~Lee, and S.-W. Seo, ``From general vision to
  reliable traversability estimation: Adapting vision foundation models for
  unstructured outdoor environments,'' 2026, arXiv:2605.29565.

\bibitem{PAPADAKIS20131373}
P.~Papadakis, ``Terrain traversability analysis methods for unmanned ground
  vehicles: A survey,'' \emph{Engineering Applications of Artificial
  Intelligence}, vol.~26, no.~4, pp. 1373--1385, 2013.

\bibitem{he2025navmoehybridmodellearningbased}
B.~He \emph{et~al.}, ``Navmoe: Hybrid model- and learning-based traversability
  estimation for local navigation via mixture of experts,'' 2025,
  arXiv:2509.12747.

\bibitem{radford2021learningtransferablevisualmodels}
A.~Radford \emph{et~al.}, ``Learning transferable visual models from natural
  language supervision,'' in \emph{Proceedings of the 38th International
  Conference on Machine Learning (ICML)}, 2021.

\bibitem{oquab2024dinov2learningrobustvisual}
M.~Oquab \emph{et~al.}, ``Dinov2: Learning robust visual features without
  supervision,'' \emph{Transactions on Machine Learning Research (TMLR)}, 2024.

\bibitem{wang2023jaccardmetriclosses}
Z.~Wang, X.~Ning, and M.~B. Blaschko, ``{J}accard metric losses: Optimizing the
  {J}accard index with soft labels,'' in \emph{Advances in Neural Information
  Processing Systems (NeurIPS)}, 2023.

\bibitem{wang2023dicesemimetriclosses}
Z.~Wang, T.~Popordanoska, J.~Bertels, R.~Lemmens, and M.~B. Blaschko, ``{D}ice
  semimetric losses: Optimizing the {D}ice score with soft labels,'' in
  \emph{Medical Image Computing and Computer Assisted Intervention (MICCAI)},
  2023.

\bibitem{lin2018focallossdenseobject}
T.-Y. Lin, P.~Goyal, R.~Girshick, K.~He, and P.~Dollár, ``Focal loss for dense
  object detection,'' \emph{IEEE Transactions on Pattern Analysis and Machine
  Intelligence}, 2020.

\bibitem{patel2025tartangroundlargescaledatasetground}
M.~Patel \emph{et~al.}, ``Tartanground: A large-scale dataset for ground robot
  perception and navigation,'' in \emph{IEEE/RSJ International Conference on
  Intelligent Robots and Systems (IROS)}, 2025.

\end{thebibliography}

\end{document}